\documentclass{article}
\usepackage{graphicx} % Required for inserting images
\usepackage{amsmath}
\usepackage{amssymb}
\usepackage{booktabs}

\title{Exploring Knowledge Transfer in Evolutionary Many-task Optimization: A Complex Network Perspective}
\author{Yudong Yang, Kai Wu, Xiangyi Teng, Handing Wang, He Yu, Jing Liu}
\date{February 2024}

\begin{document}

\maketitle
\begin{abstract}
    The field of evolutionary many-task optimization (EMaTO) is increasingly recognized for its ability to streamline the resolution of optimization challenges with repetitive characteristics, thereby conserving computational resources. This paper tackles the challenge of crafting efficient knowledge transfer mechanisms within EMaTO, a task complicated by the computational demands of individual task evaluations. We introduce a novel framework that employs a complex network to comprehensively analyze the dynamics of knowledge transfer between tasks within EMaTO. By extracting and scrutinizing the knowledge transfer network from existing EMaTO algorithms, we evaluate the influence of network modifications on overall algorithmic efficacy. Our findings indicate that these networks are diverse, displaying community-structured directed graph characteristics, with their network density adapting to different task sets. This research underscores the viability of integrating complex network concepts into EMaTO to refine knowledge transfer processes, paving the way for future advancements in the domain.
\end{abstract}

\section{Introduction}
In product design and functional implementation, there is a frequent necessity to solve analogous problems using black-box optimization models. This scenario often leads to increased computational demands and process delays. Addressing this, Evolutionary Many-task Optimization (EMaTO) has emerged as a prominent field, drawing inspiration from human cognitive abilities to leverage past experiences for new challenges \cite{5chen2019adaptive}. EMaTO research has produced a variety of algorithms, each tailored to solve problems involving analogous computational tasks. Applications of these algorithms span diverse areas, including path computation in multi-domain networks \cite{1dinh2023adaptive}, network collaborative pruning \cite{2lei2023network}, architecture search in networks \cite{3liao2023emt}, enhancement of recommendation systems \cite{4wu2023interactive}, optimization of large-scale pre-trained model sets \cite{19choong2023jack}, humanoid fault-recovery \cite{map-mto}, and model feature selection \cite{13qu2023explicit,14li2023evolutionary}. These developments indicate the field's growing significance in optimizing complex computational processes across various domains.

EMaTO algorithms differ from traditional Evolutionary Algorithms (EA) mainly in their integration of a Knowledge Transfer (KT) component, which is based on the concepts that (1) optimization processes generate valuable knowledge and (2) knowledge acquired from one task can benefit other tasks. Crucially, a robust KT mechanism in EMaTO needs to define 'WHAT' is transferred (the content of knowledge) and 'WHO' is involved in the transfer (the entities). Recent research, despite varied definitions and formal expressions, points to task similarity as a key factor in selecting subjects for transfer, utilizing methods like KLD \cite{lucas2019tile}, MMD \cite{gretton2012kernel}, SISM  \cite{wu2023transferable}, and adaptive posterior knowledge \cite{5chen2019adaptive}. Sharing knowledge between highly similar tasks often leads to better initial optimization points, faster convergence, and aids in escaping local optima. In contrast, sharing knowledge between less similar or unrelated tasks can increase the challenge of finding optimal solutions and lead to inefficient evaluations, a phenomenon known as negative transfer.

In EMaTO, mitigating negative transfer is essential, focusing on the 'WHAT' and 'WHO' of knowledge transfer mechanisms. Knowledge transfer varies in form and method, with elite individual transfer being a prevalent explicit technique where high-performing individuals from auxiliary tasks are directly injected into the target task population. Jiang \emph{et al.} \cite{jiang2023block}  introduced an approach that segments individuals into distinct blocks prior to knowledge transfer, enabling more granular transfer and effectively reducing the risk of negative transfer due to varying problem dimensions. Implicit transfer, in contrast, involves crossbreeding individuals from different tasks for evolutionary information. Addressing adaptability challenges in these methods for tasks with diverse characteristics, Feng \emph{et al.}  \cite{11feng2018evolutionary} proposed denoising autoencoders to map relationships between different tasks' search spaces, enhancing explicit transfer. This technique aims to reduce computational resource waste and overcome population convergence issues in elite individual transfer due to varying task fitness landscapes. With advances in neural networks, more complex neural network-based methods \cite{15zhou2023wgan,16liu2023multifactorial,17li2023evolutionary} have been explored for knowledge transfer, yet these have not seen wide adoption due to minimal improvement in transfer effectiveness and increased computational demands.

Another significant development has been in methods focusing on determining the subjects of knowledge transfer. The literature identifies two categories of knowledge transfer subjects: individuals and populations \cite{tan2023knowledge}. Algorithms that employ skill factors of individuals within a population for knowledge transfer and inheritance are termed multi-factorial algorithms. In contrast, multi-population algorithms assign a subpopulation to each task and facilitate knowledge transfer through inter-subpopulation interactions. While multi-population algorithms typically incur additional computational resource consumption due to the inclusion of evaluation indicators, they often reduce negative task interaction transfer and introduce more diverse methods of knowledge transfer.

This paper presents the use of network structures to describe and construct EMaTO frameworks. The motivation behind this approach is to mitigate the expensive optimization costs of assessing task similarity in large-scale many-task optimization scenarios. Previous studies have mostly balanced between minimal additional evaluations and a higher likelihood of negative transfer. To avoid repetitive task comparisons while controlling the frequency and specificity of transfer actions, thereby reducing negative transfer and enhancing EMaTO performance, we propose reconstructing the network structure with individual tasks as nodes and transfer relationships as directed edges. This approach not only controls the interaction frequency of the entire task set but also considers the elimination of negative transfer impacts on tasks through future research on sparsification.
\section{Preliminaries}
\subsection{Formulation of Multi-Task Optimization Problems}

Multi-task optimization(MTO) refers to a class of evolutionary optimization algorithms that simultaneously optimize multiple tasks by employing strategies such as knowledge transfer and knowledge sharing, thereby improving the optimization performance of each task individually. Mathematically, a set of tasks $\mathbb{T}$ to be optimized can be described as follows:

\begin{equation}
\begin{aligned}
\mathbb{T} &= \{ T_1,T_2,\dots,T_n\} \\
s.t.\quad T_i &= \min f(X_i^{D_i})  \\
X_i &= [x_i^1,x_i^2,\dots,x_i^d] \in R_i,\,i=1,2,\dots,n
\end{aligned}
\end{equation}

where $T_i$ is the $i^{th}$ task in the set $\mathbb{T}$, $f(X_i^D)$ is the objective function of the corresponding task, $D_i$ represents the dimension of the decision space, $X_i$ denotes the solution for the $i^{th}$ task, and $R_i$ is the feasible domain for the $i^{th}$ task. In this formalized description, when the number of tasks $n$ is 3 or more, we typically refer to it as a many-task optimization problem. In other words, MTO can be described as a simplified version of MaTO, and both essentially share the same optimization philosophy and specific algorithms. In this paper, we do not intend to make too much distinction between the two.

\begin{figure*}
    \centering
    \includegraphics[width=\linewidth]{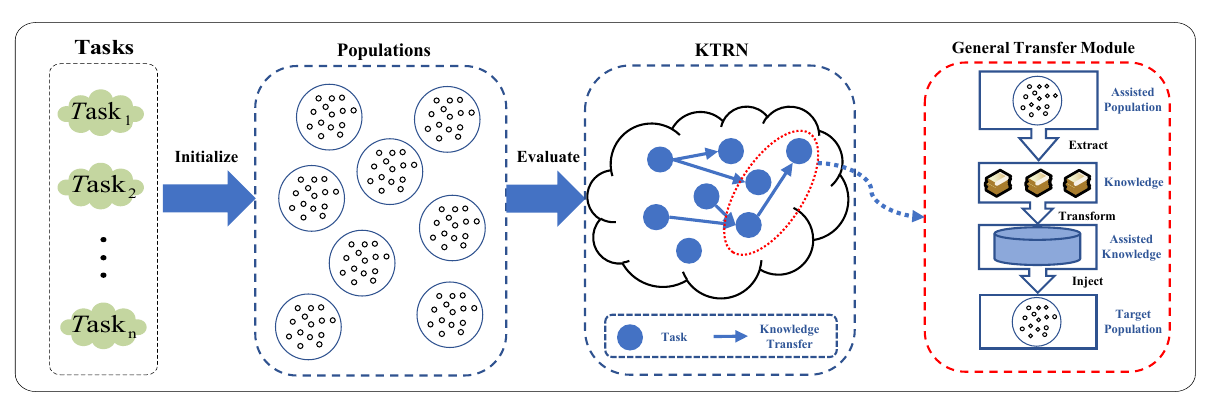}
    \caption{General process of knowledge transfer guided by KTRN in EMaTO algorithm: an illustrative diagram. This schematic represents the initialization of separate populations for each task within multi-swarm algorithms. It evaluates inter-population correlations to generate a directed network indicating knowledge transfer relationships, with each edge signifying a knowledge transfer event. The module on the right demonstrates the universal framework for knowledge transfer: extracting beneficial knowledge from an assisted population using algorithm-specific mechanisms, transforming it for the target population, and thereby injecting it to facilitate the transfer of general experience during the evolutionary process.}
    \label{fig:MTO}
\end{figure*}

\subsection{A General Framework of EMaTO}

EMaTO is an advanced form of EAs, distinguished by its capability for inter-task knowledge transfer, akin to the principles of transfer learning \cite{wang2021transfer} in deep learning. This approach leverages task similarities in terms of data or models to enhance learning efficiency on new tasks. Initially, transfer learning primarily involved using data from various tasks to bolster performance in a single task, thereby diminishing the need for extensive manual data labeling and reducing time requirements. As the field progressed, the relevance of task-related data emerged as a crucial factor in determining task similarity and guiding transfer processes.

However, unlike deep learning models where task-specific features are encapsulated in network architectures and weight configurations, EMaTO algorithms confront unique challenges. These challenges stem from their fewer adjustable parameters and lower data dependency, complicating the task comparison due to the absence of detailed task descriptors. Consequently, in the realm of EMaTO, a pivotal aspect is the extraction of relevant data from the evolutionary process to inform and direct transfer actions effectively.

In exploring transfer mechanisms within MTO, we outline its foundational structure. The seminal MTO model, the Multi-Factorial Evolutionary Algorithm (MFEA\cite{18gupta2015multifactorial}), commences with a unified population, where individuals are annotated with skill factors indicating their competency in specific tasks. This approach, however, exhibits constrained innovation in transfer dynamics, depending primarily on variations in skill factors to facilitate knowledge exchange. Recent advancements propose the initialization of distinct populations for each task, thereby enriching the diversity of transfers and augmenting the efficacy of EMaTO methods. Our discussion predominantly centers on these multi-population frameworks in EMaTO.

Figure \ref{fig:MTO} outlines the streamlined workflow for knowledge transfer in EMaTO. Each task begins with an initial population, which evolves into a repository of solutions. This repository contains either elite individuals or reflects the task's population distribution. The General Transfer Module illustrates the extraction and transformation of transferable knowledge from an assisted task's population, subsequently integrated into the target task. This process can encompass direct migration of elite individuals or the utilization of population distribution data.

\subsection{Networks}
Networks, serving as data structures to delineate relationships between entities, are integral in representing complex networks. An directed network, denoted as $G = (V, E)$, where $V$ is the set of nodes and $E$ is the set of ordered pairs of edges, such as  $(u, v)$, indicating a unidirectional relationship from node $u$ to node $v$.  %Their application spans diverse real-world domains, including social networks, academic citation networks, and electrical transmission networks.
In EMaTO, networks facilitate the understanding of intricate relationships and trends, aiding decision-making and system optimization. Central tasks in network analysis, such as link prediction \cite{lu2011link}, node clustering \cite{kumarawadu2008algorithms}, and node classification \cite{wei2018identifying}, provide insights into potential connections, groupings of similar nodes, and labeling of node attributes, respectively.

%The former is prevalently employed in contexts like social networking, whereas the latter is more pertinent to applications involving directed linkages, such as in web structures.

For EMaTO, networks effectively model knowledge transfer behaviors. Nodes represent tasks or their respective populations, while edges signify the transfer of knowledge between these tasks. Addressing the dynamic nature of task interactions, contemporary approacheslike SBO-type algorithms \cite{liaw2019evolutionary} suggest using directed networks. These capture the evolving, heterogeneous nature of task relationships more accurately than traditional undirected or unidirectional structures. This approach acknowledges the complexity and diversity of connections in evolutionary optimization, where edges in the network evolve and vary in significance. Initially, our research explores these interactions within the simplified framework of basic directed networks, given the challenges in constructing and analyzing more intricate network models.

Figure \ref{fig:MTO} illustrates networks as pivotal in facilitating knowledge transfer in EMaTO. Task correlations, derived from population evaluations using targeted metrics, guide the development of a Knowledge Transfer Relationship Network (KTRN). In KTRN, tasks are represented by vertices, and actual knowledge transfer interactions are represented by edges.
\section{Advancements in EMaTO Algorithms}

This study categorizes and evaluates evolutionary many-task algorithms, differentiating them based on their approaches to knowledge transfer: multi-factorial or multi-population types. We assess these algorithms using a unified testing suite, focusing on the effectiveness of network metrics in various evolutionary contexts.

\subsection{Algorithmic Classification and Comparison}
We classify evolutionary many-task optimization algorithms into two primary categories: multi-factorial and multi-population, each representing a distinct approach to knowledge transfer. Multi-factorial algorithms integrate multiple tasks within a single population, distinguishing individuals by skill factors. In contrast, multi-population algorithms allocate distinct populations or optimizers to each task, promoting knowledge reuse and collaborative evolution through inter-population transfers.

\subsubsection{Multi-factorial Approach}
Initially, multi-factorial algorithms addressed multi-task challenges within a single population framework. These algorithms segment sub-populations by assigning skill factors to individuals, facilitating task-specific optimization through traditional crossover mechanisms. The process typically involves evaluating individual performance across tasks, generating offspring through crossover or mutation, and transferring skill factors across generations. A notable aspect of multi-factorial algorithms like MFEA is the use of a random mutation probability ($rmp$) parameter to regulate network link formation, although some variants propose adaptive rmp adjustments.

\subsubsection{Multi-Population Approach}
In contrast, multi-population algorithms create separate sub-populations for each task, enabling both independent and assisted evolutionary states. Independent evolution mirrors single-task evolutionary algorithms, while assisted evolution incorporates external knowledge, such as elite individuals or evolutionary parameters, into the evolutionary process. These algorithms are characterized by their collaborative evolution mechanisms and involve strategic knowledge transfer guided by specific metrics and performance evaluations. This approach includes independently evolving each population, assessing inter-task transfer probabilities, and systematically executing knowledge transfer based on metric-driven task relationships.
\subsection{Algorithm Overview}
In this study, we comprehensively compare various evolutionary algorithms, including MFEA \cite{18gupta2015multifactorial}, BoKTDE \cite{7jiang2022bi}, EMaTO-MKT \cite{liang2021evolutionary}, MTEA-AD \cite{8wang2021solving}, MaTDE \cite{5chen2019adaptive}, and SBCMAES \cite{21liaw2017evolutionary}. These algorithms each have their unique features, showcasing the diversity and innovation in the field of evolutionary computation. For instance, MFEA facilitates knowledge transfer through cross-task genotypic representation, emphasizing the similarity and interdependence between tasks; BoKTDE employs a dual-objective metric to select different knowledge transfer strategies; EMaTO-MKT employs unsupervised clustering techniques to categorize tasks into groups, ensuring that all knowledge transfer occurs within these groups to mitigate the risk of negative transfer; MTEA-AD identifies and utilizes beneficial knowledge transfers through anomaly detection; MaTDE uses an archive mechanism combined with reward and punishment systems to control whether transfer occurs between individual tasks; and SBCMAES adopts an evolutionary strategy based on adaptive covariance matrix as its base optimizer, utilizing characteristics of symbiosis in biocoenosis to manage the probability of knowledge transfer. These algorithms provide a wide range of solutions for different optimization problems and challenges, reflecting the adaptability and diversity of evolutionary optimization strategies in handling complex many-task problems.

In the widely influential multi-factorial evolutionary algorithm MFEA \cite{18gupta2015multifactorial}, the interaction frequency between different tasks is determined by an external parameter, random mutation probability (rmp). Under this mechanism, although individuals in the population use skill factors to label their "tasks of expertise," the overall interaction frequency in the task set shows a similar pattern regardless of the actual relationships between tasks, which is unreasonable for tasks with insufficient similarity to warrant borrowing, and can even exacerbate negative transfer phenomena, much like how the coordination methods and tactics of a top-tier soccer team may not be suitable for another baseball team. Clearly, Gupta \emph{et al.}  \cite{20bali2019multifactorial} quickly recognized the insensitivity of the MFEA algorithm to negative transfer and its research potential in this direction, proposing the MFEA-II algorithm, which uses an adaptively changing RMP matrix to adjust the probability of interaction between tasks. In extreme cases, when the learned RMP matrix is all zeros, it is considered as a set of parallel single-task EAs, meaning no inter-task transfer occurs.

Multi-population algorithms have a more diverse range of transfer methods and judgment criteria. Rung-Tzuo Liaw \emph{et al.}  \cite{21liaw2017evolutionary} argued that the relationship between tasks is not just similar or dissimilar, but can also be learning from each other, competing, or parasitic, among others. Based on Symbiosis in Biocoenosis, the SBO algorithms adaptively adjust the probability of transfer behavior based on the improvement caused by inter-population knowledge transfer, effectively categorizing the relationships between tasks into six types, and conducting varying degrees of knowledge transfer or no transfer actions within different categories. EMaTO-MKT \cite{liang2021evolutionary} uses unsupervised k-means clustering to divide the entire task set into different clusters guided by prior knowledge, then selects a certain number of assisted tasks within the same cluster for each task during the evolutionary process to complete knowledge transfer.

The aforementioned methods primarily simplify task selection for knowledge transfer by categorizing tasks into various subsets. Many algorithms are dedicated to developing different methods of assessing task similarity, such as MaTEA \cite{5chen2019adaptive}, which uses KLD as an important indicator to summarize the overall state of populations based on the differences in probability distributions between them. It also uses a scaling factor $\lambda$ to prevent negative transfer phenomena in cases of similar distributions. These mechanisms, compared to MFEA-type algorithms, take into account the population differences caused by task characteristics when choosing transfer targets. MTEA-AD \cite{8wang2021solving} uses population probability distribution to capture task similarities and dynamically selects candidate transfer individuals from normal data using a density-based anomaly detection model, thereby reducing the risk of negative transfer. The BoKT framework \cite{7jiang2022bi} proposes using dual-objective similarity metrics to assess both Shape similarity and Domain similarity to describe the global and optimum relationships between tasks, binding the two indicators with different knowledge transfer strategies. This allows for adaptive changes in cooperative strategies between tasks based on specific needs, resulting in significant improvements in algorithm performance.
\subsection{Experiments}
\subsubsection{Benchmark}
To describe the task transfer networks generated during the computation process of different algorithms, we have chosen the single-objective MATOPs from WCCI-20 as the testing benchmark to evaluate the performance of each algorithm and to extract the task transfer networks constructed during their operation. The WCCI-20 single-objective MaTOPs benchmark consists of 5 MaTOPs, each comprising 50 single-objective optimization tasks. These tasks are derived by translating and rotating basic optimization functions such as Rosenbrock, Ackley, Rastrigin, Griewank, Weierstrass, and Schwefel, resulting in task sets with complex and diverse structures. The P1-P3 task sets consist only of the same basic optimization function, while the other task sets contain three or more base optimization functions, increasing the diversity between tasks and the difficulty in judging the need for knowledge transfer. The different base optimization functions possess distinct optimization landscape objectives, providing an opportunity to test the transfer methods of algorithms in various scenarios. Furthermore, each MaTOP internally uses only the same base optimization function, which aligns more closely with real-world scenarios of parallel computation of similar tasks.
\begin{table}[!h]
    \centering
    \caption{Basic task composition of the WCCI-20 benchmark}
    \resizebox{\textwidth}{!}{
    \label{table:benchmarks}
    \begin{tabular}{cccccccc}
    \toprule
    \textbf{} & \textbf{Sphere} & \textbf{Ackley} & \textbf{Rosenbrock} & \textbf{Rastrigin} & \textbf{Griewank} & \textbf{Weierstrass} & \textbf{Schwefel} \\ \midrule
    \textbf{P1} & \checkmark & ~ & ~ & ~ & ~ & ~ & ~ \\ 
    \textbf{P2} & ~ & ~ & \checkmark & ~ & ~ & ~ & ~ \\ 
    \textbf{P3} & ~ & \checkmark & ~ & ~ & ~ & ~ & ~ \\ 
    \textbf{P4} & \checkmark & ~ & \checkmark & \checkmark & ~ & ~ & ~ \\ 
    \textbf{P5} & ~ & ~ & ~ & \checkmark & \checkmark & ~ & ~ \\ 
    \textbf{P6} & ~ & \checkmark & ~ & ~ & ~ & \checkmark & \checkmark \\ 
    \textbf{P7} & ~ & \checkmark & \checkmark & \checkmark & \checkmark & ~ & ~ \\ 
    \textbf{P8} & ~ & ~ & \checkmark & \checkmark & \checkmark & \checkmark & ~ \\ 
    \textbf{P9} & ~ & \checkmark & ~ & \checkmark & \checkmark & \checkmark & \checkmark \\ 
    \textbf{P10} & \checkmark & ~ & ~ & \checkmark & ~ & \checkmark & \checkmark \\ \bottomrule
    \end{tabular}
    }
\end{table}

\subsubsection{Optimization Performance Comparison}
\begin{figure*}[!ht]
    \centering
    \includegraphics[width=1\linewidth]{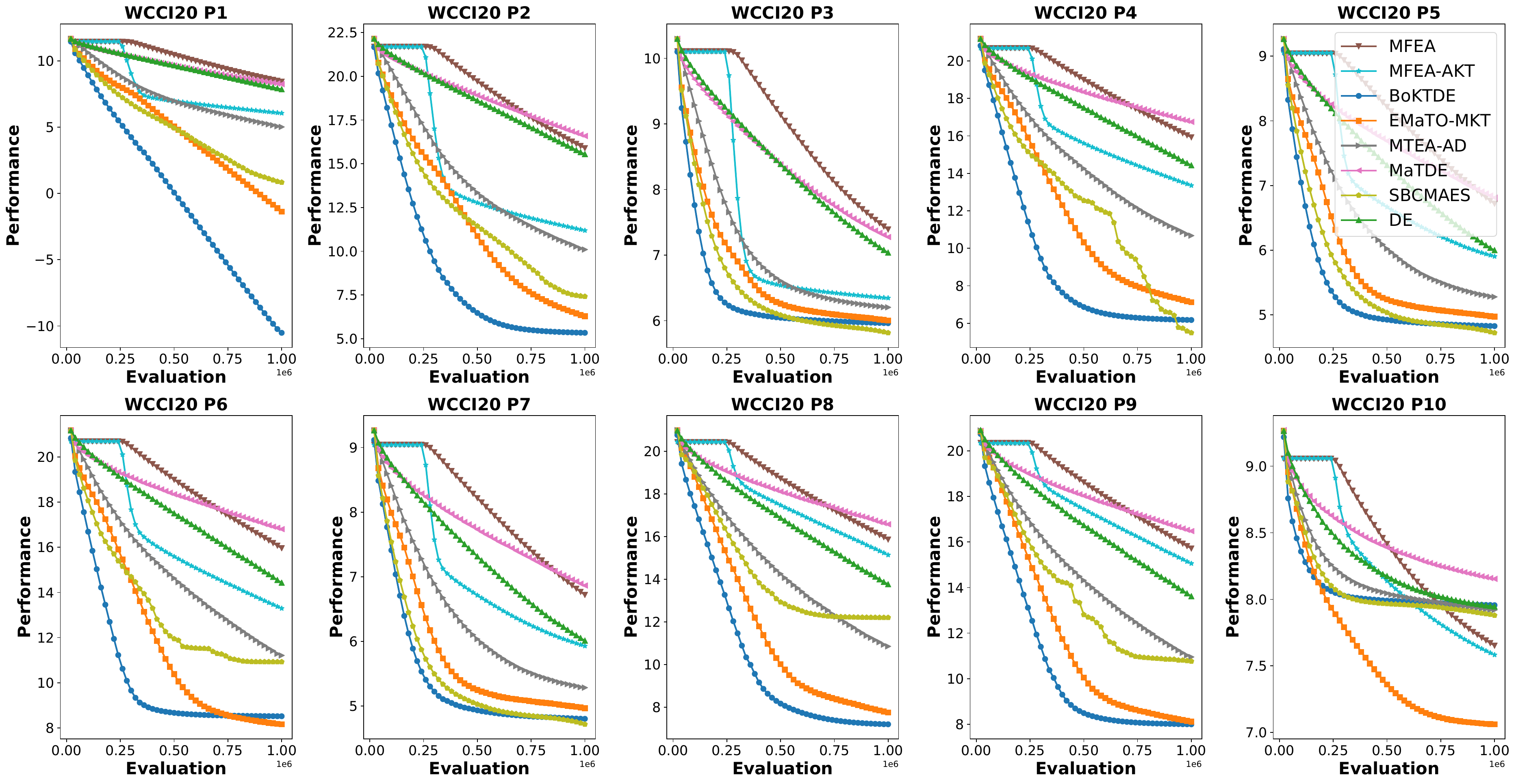}
    \caption{Convergence performance comparison of six renowned EMaTO algorithms versus single-task DE algorithm on the WCCI-20 MaTOP benchmark}
    \label{fig:performance-curves}
\end{figure*}

Under the same evaluation iteration limit and using parameters given in the corresponding literature, we conducted numerical optimization of different algorithms on task sets and collected the experimental results. The results are compiled in Table \ref{algos_ability}, where each row represents a task set from WCCI20 and each column represents a specific algorithm tested. The algorithm's effectiveness is represented by the average optimization value achieved in the corresponding task set. The data in the table indicates that BoKTDE, EMaTO-MKT, and SBCMAES demonstrate superior optimization performance in most task sets, suggesting that these types of algorithms excel in the process of knowledge transfer compared to others. For algorithms belonging to different categories, multi-population algorithms show significantly better results than multi-factorial algorithms under the same task set and evaluation iteration constraints, especially in P1, P2, P4, and P6, where multi-population algorithms lead by no less than three orders of magnitude on average. Additionally, we included the traditional single-task algorithm Differential Evolution in the algorithm convergence performance comparison chart.As shown in Fig\ref{fig:performance-curves}, it's evident that EMaTO algorithms exhibit better convergence speed in the early stages of evolution. This is because the transfer conducted in the early stages rapidly enhances the exploratory capabilities of tasks within the task set, far exceeding those during single-task evolution. Indeed, the capability of EMaTO lies in the use of generalized knowledge to provide approaches for solving similar problems, whether it involves providing better initial solutions or enhanced exploratory capabilities to escape local optima.

\begin{table}[h]
    \centering
    \caption{Results of six algorithm conducted on WCCI20}
    \resizebox{\textwidth}{!}{ % Resizes the table to fit within the text width
    \begin{tabular}{cccccccc}
    \toprule
        \textbf{} & \textbf{MFEA} & \textbf{BoKTDE} & \textbf{EMaTO-MKT} & \textbf{MTEA-AD} & \textbf{MaTDE} & \textbf{SBCMAES} \\ \midrule
    \textbf{P1} & 4.515e+03 & 2.428e-05 & 2.739e-01 & 1.535e+02 & 3.798e+03 & 1.624e+00 \\ 
    \textbf{P2} & 8.655e+06 & 1.748e+02 & 6.094e+02 & 2.354e+04 & 1.584e+07 & 2.640e+02 \\ 
    \textbf{P3} & 1.647e+03 & 3.885e+02 & 4.039e+02 & 4.941e+02 & 1.437e+03 & 3.339e+02 \\ 
    \textbf{P4} & 8.082e+06 & 5.705e+02 & 1.353e+03 & 4.519e+04 & 1.857e+07 & 2.630e+02 \\ 
    \textbf{P5} & 8.250e+02 & 1.253e+02 & 1.436e+02 & 1.960e+02 & 9.101e+02 & 1.140e+02 \\ 
    \textbf{P6} & 8.876e+06 & 4.968e+03 & 3.447e+03 & 7.730e+04 & 1.981e+07 & 2.305e+04 \\ 
    \textbf{P7} & 8.256e+02 & 1.219e+02 & 1.444e+02 & 1.971e+02 & 9.565e+02 & 1.115e+02 \\ 
    \textbf{P8} & 7.677e+06 & 1.595e+03 & 3.093e+03 & 5.325e+04 & 1.553e+07 & 8.539e+04 \\ 
    \textbf{P9} & 7.010e+06 & 2.994e+03 & 2.575e+03 & 5.570e+04 & 1.423e+07 & 1.067e+05 \\ 
    \textbf{P10} & 2.114e+03 & 2.852e+03 & 1.160e+03 & 2.809e+03 & 3.482e+03 & 2.607e+03 \\ \bottomrule
    \end{tabular}
    }
    \label{algos_ability}
\end{table}

\section{Evaluating KTRN in EMaTO}
This section discusses the formation, significance, and effective metrics for knowledge transfer networks in evolutionary many-task optimization, particularly focusing on the networks generated by exemplary algorithms. We analyze task interactions within these networks from a complex network perspective.

\subsection{Capturing KTRN in Algorithm Execution}
The knowledge transfer network in the execution of Evolutionary Many-Task Optimization algorithms, often not explicitly stored, presents unexplored characteristics from a network view. This study introduces a novel approach to systematically analyze transfer relationships between task sets during algorithm execution, providing insights into knowledge transfer behaviors in EMaTO algorithms.

We modified algorithms to track all knowledge transfers, recording source and target task pairs in an \(N \times N\) adjacency matrix after each evolutionary cycle. For multi-factorial algorithms, we determine the occurrence of knowledge transfer relationships through vertical cultural inheritance in the algorithm framework. That is, if the offspring is mutated from a single individual or obtained by crossing two individuals with the same skill factor, then the transfer relationship does not occur. Only when parent individuals possess different skill factors do we consider that the task corresponding to the skill factor inherited by the offspring has accepted assistance from another task in the parent individuals. For multi-population algorithms, this process is relatively simple, as algorithms typically have a separate knowledge transfer module to handle knowledge transfers, allowing us to record the participating source and target tasks in the corresponding positions in the adjacency matrix during each actual transfer. Despite the differences in whether transfers occur between tasks and the specific transfer methods in different multi-population algorithms, our recording method requires little change.

\subsection{Metrics for Evaluating KTRN}
We collected the adjacency matrices of actual knowledge transfer relationships occurring in each generation when solving the WCCI20 MaTOP with different algorithms under the default parameters given in the literature. After converting these adjacency matrices into directed networks, we calculated several evaluation metrics for the networks, recorded in Table \ref{tab:structural_metrics}. Each row in the table represents the metrics of all networks generated by an algorithm across all task sets, while each column represents a specific metric value, showcasing different characteristics of the networks from various perspectives.

\subsubsection{Overall Network Properties}
\begin{table*}[h]
    \centering
    \caption{KTRN metric - overall structural properties}
    \resizebox{\textwidth}{!}{
    \begin{tabular}{ccccccc}
    \toprule
        & \textbf{BOKTDE} & \textbf{EMATO-MKT} & \textbf{MFEA} & \textbf{MTEA-AD} & \textbf{MaTDE} & \textbf{SBCMAES} \\ \midrule
    \textbf{D} & 0.020 (0.000) & 0.102 (0.000) & 0.453 (0.013) & 0.021 (0.009) & 0.002 (0.001) & 0.010 (0.001) \\ 
    \textbf{C} & 0.003 (0.007) & 0.506 (0.056) & 0.449 (0.014) & 0.049 (0.037) & 0.001 (0.001) & 0.001 (0.000) \\ 
    \textbf{DIA} & \textasciitilde & \textasciitilde & 2.007 (0.084) & \textasciitilde & \textasciitilde & \textasciitilde \\ \bottomrule
    \end{tabular}
    }
    \label{tab:structural_metrics}
\end{table*}

\paragraph{\textbf{Density}}
Network density, a crucial metric in network analysis, measures the closeness of a network to a complete network. It's defined by the ratio of existing edges to the maximum possible edges among vertices:
\begin{equation}
    D = \frac{2|E|}{|V|(|V|-1)}
\end{equation}
where \(|E|\) is the edge count, and \(|V|\) is the vertex count. In knowledge transfer networks, this relates directly to the frequency of knowledge transfer actions, with edges representing transfers and vertices representing optimization tasks. A higher network density suggests more frequent knowledge transfer, indicating an intensive interaction among tasks. Our analysis shows varied densities across different algorithms, with notable examples being BOKTDE, EMATO-MKT, and SBCMAES exhibiting densities of 0.0204, 0.1020, and 0.0104, respectively. This variation reflects the differing extent of knowledge transfer activities among these algorithms, with MFEA showing a distinct pattern compared to others. Such insights are critical for understanding and optimizing knowledge transfer strategies in evolutionary optimization.

\paragraph{\textbf{Diameter}}
The network diameter is the longest shortest path between any two nodes in the network. It provides a measure of the network's linear size and is defined as:
\begin{equation}
    DIA = \max_{i,j} d(i,j),
\end{equation}
where $( d(i,j) )$  is the shortest path distance between nodes $( i )$  and $( j )$. In complex networks, the diameter is used to evaluate the connectivity and compactness of the overall network structure. A smaller value typically means that there is always a relatively short path connecting different nodes in the entire network. In other words, a smaller diameter usually indicates that the network possesses certain small-world properties. Among the algorithms tested, only the network corresponding to MFEA had an average diameter of 2.0072, while the others were not applicable, indicating that the KTRN have fragmented characteristics, proving that different evaluation methods can lead to the same conclusion: knowledge transfer between tasks is not always effective across all tasks but is usually limited to a small number of local neighbor tasks.

\paragraph{\textbf{Clustering Coefficient}}
The clustering coefficient is a measure of the extent to which nodes in a network tend to cluster together. It is defined for each node as the ratio of the actual number of links between its neighbors to the maximum possible number of links, reflecting the local density of connected triangles. The clustering coefficient for the entire network is the average of these coefficients across all nodes, represented as follows:
\begin{align}
    C &= \frac{1}{n}\sum_{i=1}^{n}C_i, \\
    C_i &= \frac{2T_i}{deg_i(deg_i-1)}
\end{align}
where $C$ is the average clustering coefficient of the network, $C_i$ is the clustering coefficient of the $i$th node, $T_i$ represents the number of triangles connected to node $i$, and $deg_i$ is the degree of node $i$, or the number of edges connected to it. The factor $n$ denotes the total number of nodes in the network.

In our analysis, EMATO-MKT and MFEA demonstrated significant clustering, with average coefficients of 0.5062 and 0.4486, respectively. This observation is attributed to EMATO-MKT’s incorporation of unsupervised k-means clustering to evaluate task similarity. Networks with lower edge densities exhibited less pronounced clustering, indicating a correlation between network density and clustering tendencies.

\subsubsection{Local Connectivity Properties}

\begin{table*}[h]
    \centering
    \caption{KTRN metrics - local connectivity properties}
    \resizebox{\textwidth}{!}{
    \begin{tabular}{ccccccc}
    \toprule
        & \textbf{BOKTDE} & \textbf{EMATO-MKT} & \textbf{MFEA} & \textbf{MTEA-AD} & \textbf{MaTDE} & \textbf{SBCMAES} \\ \midrule
    \textbf{A} & 0.0(0.0) & 0.0(0.0) & -0.454 (0.353) & 0.0(0.0) & 0.0(0.0) & 0.0(0.0) \\ \hline
    \textbf{SAC} & 0.191 (0.182) & 0.131 (0.059) & 0.453 (0.013) & 0.003 (0.002) & 0.046 (0.021) & 0.061 (0.031) \\ \hline
    \textbf{H} & 0.988 (0.304) & 0.268 (0.087) & 0.137 (0.018) & 1.650 (0.520) & 2.308 (0.710) & 1.625 (0.314) \\ \bottomrule
    \end{tabular}
    }
    \label{tab:connectivity_metrics}
\end{table*}

\paragraph{\textbf{Assortativity}}
Assortativity in networks quantifies the propensity of nodes to connect with others that are similar in certain aspects, particularly in terms of node degree. The assortativity coefficient \(A\) is calculated using the formula:
\begin{equation}
    A = \frac{\sum_{ij}(A_{ij}-\frac{k_i k_j}{2m})k_i k_j}{\sum_{ij}(k_i \delta_{ij}-\frac{k_i k_j}{2m})k_i k_j},
\end{equation}
where \(A_{ij}\) is the adjacency matrix, \(k_i\) and \(k_j\) are the degrees of nodes \(i\) and \(j\) respectively, \(m\) is the total number of edges, and \(\delta_{ij}\) is the Kronecker delta function.

Our analysis revealed that the MFEA algorithm displayed a unique negative assortativity pattern, indicating a preference for knowledge transfer between tasks with dissimilar degrees of existing relationships. In contrast, other algorithms showed no significant assortativity tendencies. This suggests that unlike MFEA, which adjusts transfer based on node connections through random mutation probability (rmp) control, other algorithms do not explicitly link transfer behavior to the current degree of nodes. This distinction highlights the diverse approaches to establishing knowledge transfer relationships in evolutionary optimization algorithms.

\paragraph{\textbf{Subgraph Average Connectivity}}
Subgraph Average Connectivity (SAC) is a crucial metric for understanding the internal structure of knowledge transfer networks, which often consist of several disconnected subgraphs. SAC measures the internal connection density within these subgraphs and is defined as:
\begin{equation}
SAC = \frac{1}{m} \sum_{i=1}^{m} \frac{|E_i|}{\delta_i(\delta_i-1)}
\end{equation}
where $m$ represents the number of subgraphs, $$|E_i|$$ the number of edges in the $i$th subgraph, and $\delta_i$ the number of nodes in it. This metric indicates the average likelihood of knowledge transfer between tasks within a subgraph. 

In our analysis, we observed that the MFEA algorithm's entire network is connected, making its SAC equal to its network density. Conversely, for algorithms like SBCMAES, the SAC value significantly exceeds the network's average density, implying a higher concentration of connections within subgraphs. This pattern suggests the presence of community structures within the knowledge transfer network. Notably, algorithms employing clustering in task evaluation, like EMATO-MKT, exhibit more uniform internal connection densities across their network structures, reflecting a balanced distribution of connections within subgraphs.

\paragraph{\textbf{Heterogeneity}}
Network heterogeneity measures the variation in node degree distribution, a key aspect in understanding network structures. It's calculated using the formula:
\begin{equation}
    H = \frac{\sigma}{\langle k \rangle}
\end{equation}
where $\sigma$ represents the standard deviation of the node degrees, and $\langle k \rangle$ is the average degree of the network. High heterogeneity indicates a significant variance in node connectivity, suggesting small-world characteristics in knowledge transfer networks. This could mean that some nodes (hubs) possess higher entropy knowledge suitable for broader transfer, in contrast to the majority with less transferable knowledge. Our findings show heterogeneity in most networks generated by the tested algorithms, except for MFEA, with its random edge generation, and EMATO-MKT, which standardizes node degrees through clustering. Such heterogeneity underlines the presence of influential hub nodes, emphasizing the need for nuanced task evaluation and knowledge transfer methodologies.

\section{Adjusting KTRN for Improved Transfer Effects}
In previous sections, we explored the performance of existing algorithms under default parameter settings and extracted the knowledge transfer relationship networks from the algorithm's execution process, analyzing their characteristics in conjunction with the principles of the algorithms. In this chapter, we attempt to validate the actual impact of the knowledge transfer relationship network on algorithm performance by adjusting the network's characteristics.

\begin{table*}[!t]
    \centering
    \caption{Network metrics of knowledge transfer relation networks in EMaTO-MKT algorithm under various parameter settings for solving the WCCI 20 MaTOP benchmark}
    \label{yab:mkt-para-metric}
    \resizebox{\textwidth}{!}{
    \begin{tabular}{cccccccccc}
    \toprule
        \textbf{} & \multicolumn{3}{c}{$K$ = 3}  & \multicolumn{3}{c}{$K = 5$}  & \multicolumn{3}{c}{$K = 10$}  \\ \midrule
        \textbf{} & $N=3$ & $N=5$ & $N=10$ & $N=3$ & $N=5$ & $N=10$ & $N=3$ & $N=5$ & $N=10$  \\ \midrule
        \textbf{D} & 0.0612 (0.0) & 0.102 (0.0) & 0.2041 (0.0) & 0.0612 (0.0) & 0.102 (0.0) & 0.2041 (0.0) & 0.0612 (0.0) & 0.102 (0.0) & 0.2041 (0.0)  \\ 
        \textbf{C} & 0.396 (0.088) & 0.4575 (0.079) & 0.566 (0.079) & 0.402 (0.083) & 0.470 (0.079) & 0.570 (0.083) & 0.404 (0.084) & 0.478 (0.077) & 0.583 (0.083)  \\ 
        \textbf{H} & 0.459 (0.184) & 0.415 (0.141) & 0.338 (0.101) & 0.418 (0.189) & 0.375 (0.154) & 0.323 (0.100) & 0.422 (0.193) & 0.353 (0.159) & 0.301 (0.109)  \\ 
        \textbf{SAC} & 0.113 (0.096) & 0.146 (0.074) & 0.240 (0.094) & 0.103 (0.067) & 0.143 (0.071) & 0.248 (0.103) & 0.120 (0.091) & 0.130 (0.068) & 0.237 (0.091)  \\ \bottomrule
    \end{tabular}
    }
\end{table*}

\begin{figure}[]
    \centering
    \includegraphics[width=\linewidth]{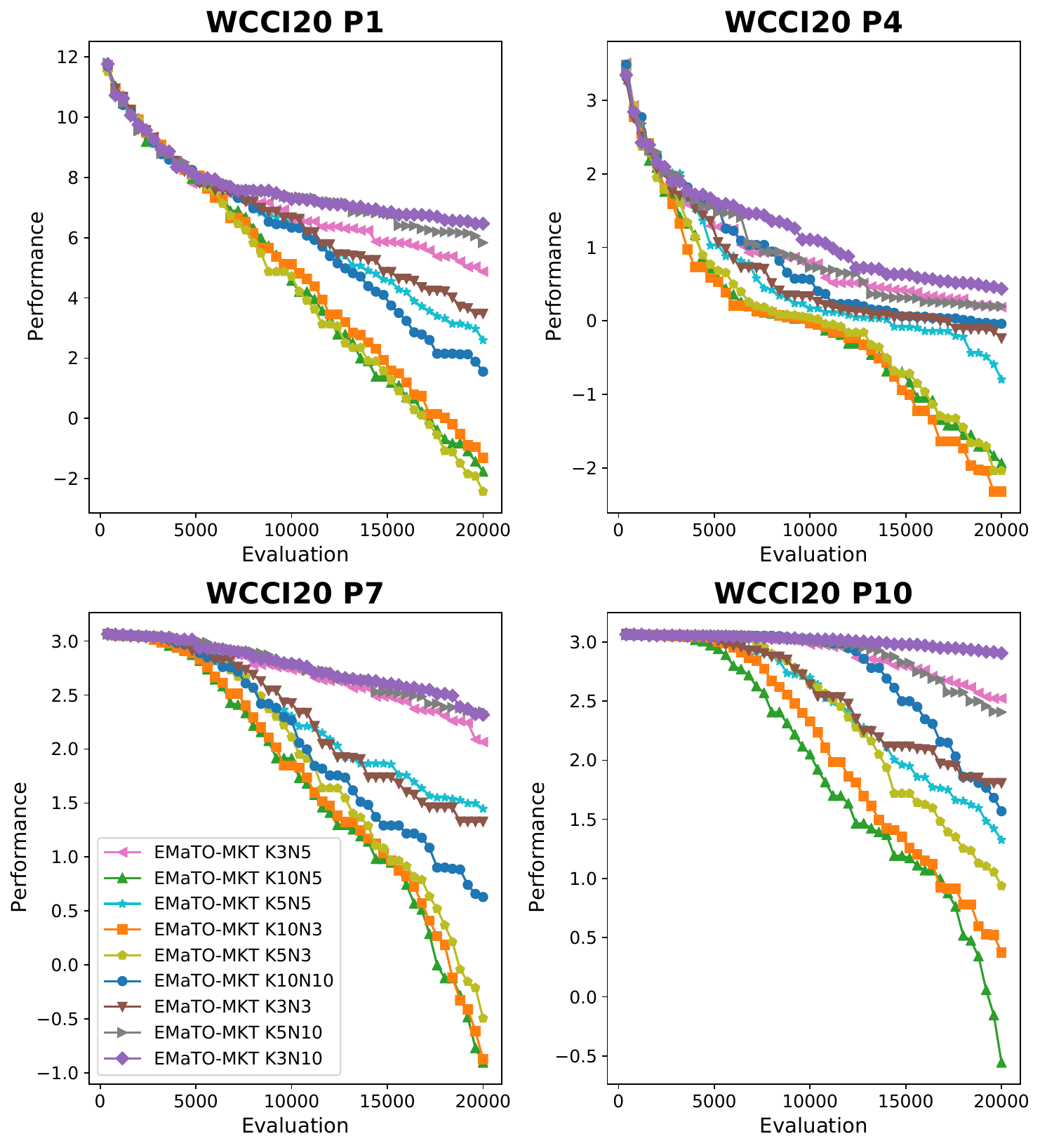}
    \caption{Convergence performance comparison of the EMaTO-MKT algorithm on select tasks from the WCCI20 MaTOP benchmark under various parameter combinations.}
    \label{fig:param_adjust}
\end{figure}

For most EMaTO algorithms, the capability to adjust the frequency of knowledge transfer between different tasks through the setting of hyperparameters is often a default feature. Hence, we can indirectly explore the impact of network density on knowledge transfer effectiveness by adjusting these parameters. Among existing algorithms, EMaTO-MKT has the most hyperparameters that can be used to adjust the network structure generated during algorithm execution. Moreover, due to the use of clustering algorithms in its framework, EMaTO-MKT can stably generate networks with structural similarities, providing good support for our tests. In this section, we choose to adjust two parameters in the algorithm, $K$ and $N$, to control the knowledge transfer relationship network, where $K$ represents the number of task clusters obtained through the k-means algorithm during the clustering process, and tasks within these subsets are considered potential auxiliary tasks for each other. $N$ represents the number of tasks that can assist in their own evolution within the current task's subset in one iteration. By adjusting these two parameters, we can preliminarily divide communities and adjust network density across the entire task set. For this part of the experiment, we selected 9 pairs of parameter settings for exploration, namely combinations of $K=3, 5, 10$ and $N=3, 5, 10$. After conducting 30 parallel experiments using the WCCI20 MaTOP benchmark across all parameter combinations, we obtained the performance of each parameter combination. We present some results in Figure \ref{fig:param_adjust} and also collected the relationship networks of knowledge transfer interactions between tasks during algorithm execution, displaying the metric values of networks corresponding to different parameters in Table \ref{yab:mkt-para-metric}.

\begin{table}[!t]
    \centering
    \caption{Comparison of best and worst performance counts of EMaTO-MKT algorithm under different parameter settings in WCCI 20 MaTOP benchmark}
    \label{table:MKT2Para}
    \resizebox{\textwidth}{!}{
    \begin{tabular}{cccccccccc}
    \toprule
        \textbf{} & \multicolumn{3}{ c }{$K = 3$}  & \multicolumn{3}{ c }{$K = 5$}  & \multicolumn{3}{ c }{$K = 10$}  \\ \midrule
        \textbf{} & $N=3$ & $N=5$ & $N=10$ & $N=3$ & $N=5$ & $N=10$ & $N=3$ & $N=5$ & $N=10$  \\ \midrule
        \textbf{P1} & 0 - 0 & 0 - 0 & 0 - 48 & 14 - 0 & 0 - 0 & 0 - 2 & 22 - 0 & 14 - 0 & 0 - 0  \\ 
        \textbf{P2} & 1 - 0 & 0 - 0 & 0 - 45 & 11 - 0 & 0 - 0 & 0 - 5 & 20 - 0 & 18 - 0 & 0 - 0  \\ 
        \textbf{P3} & 1 - 0 & 0 - 0 & 0 - 37 & 18 - 0 & 3 - 0 & 0 - 13 & 15 - 0 & 9 - 0 & 4 - 0  \\ 
        \textbf{P4} & 0 - 0 & 0 - 0 & 0 - 47 & 12 - 0 & 0 - 0 & 0 - 3 & 23 - 0 & 15 - 0 & 0 - 0  \\ 
        \textbf{P5} & 1 - 2 & 0 - 0 & 0 - 38 & 13 - 0 & 1 - 0 & 0 - 10 & 14 - 0 & 18 - 0 & 3 - 0  \\ 
        \textbf{P6} & 2 - 0 & 0 - 0 & 0 - 36 & 12 - 0 & 1 - 2 & 1 - 8 & 18 - 3 & 16 - 0 & 1 - 1  \\ 
        \textbf{P7} & 0 - 1 & 0 - 1 & 0 - 43 & 14 - 0 & 2 - 0 & 0 - 6 & 13 - 0 & 20 - 0 & 1 - 0  \\ 
        \textbf{P8} & 0 - 0 & 0 - 1 & 0 - 44 & 15 - 0 & 2 - 1 & 0 - 5 & 18 - 0 & 13 - 0 & 2 - 1  \\ 
        \textbf{P9} & 1 - 2 & 0 - 0 & 0 - 42 & 11 - 1 & 1 - 0 & 0 - 4 & 16 - 0 & 16 - 0 & 5 - 1  \\ 
        \textbf{P10} & 0 - 0 & 0 - 0 & 0 - 39 & 11 - 1 & 4 - 1 & 0 - 8 & 15 - 1 & 13 - 0 & 7 - 0  \\ \midrule
        \textbf{Total} & \textbf{6 - 5} & \textbf{0 - 2} & \textbf{0 - 395} & \textbf{130 - 2} & \textbf{13 - 2} & \textbf{1 - 57} & \textbf{174 - 3} & \textbf{152 - 0} & \textbf{23 - 1}  \\ \bottomrule
    \end{tabular}
    }
\end{table}

In selected algorithms, the parameter $N$ controls the number of auxiliary tasks for each task during the knowledge transfer process, so we can essentially use the parameter $N$ to directly correspond to a fixed network density value, as shown in Table \ref{yab:mkt-para-metric}. In Table \ref{table:MKT2Para}, we recorded the number of times each parameter combination achieved the best and worst optimization performance among the 50 independent tasks in the current task set, where the left side represents the number of times it was the best, and the right side represents the number of times it was the worst. Among the nine parameter combinations, the combinations $(K = 5, N = 3)$, $(K = 10, N = 3)$, and $(K= 10, N = 5)$ all achieved good results, while all combinations with $K = 3$ exhibited the worst optimization performance compared to others. This is due to the lack of sufficient community division of tasks, leading to so-called negative transfer between irrelevant tasks. Moreover, all parameter combinations corresponding to $N = 10$ also achieved poorer optimization results because a large number of auxiliary tasks made it easy to select nearly irrelevant tasks even within the same community. Conversely, when the task clusters are divided finely enough, all tasks within a community will have a relatively high level of relatedness.

Apart from this, Figure \ref{fig:param_adjust} reveals that, although algorithm performance on task sets under varied parameters initially lacks significant variance, the differences escalate with more evaluations. Initially, random exploration in evolutionary algorithm optimization can incidentally improve performance, allowing even irrelevant knowledge transfer to benefit the current task. As optimization advances, the necessity for precise, effective knowledge grows to overcome local optima. During this phase, improper community segmentation of task sets or overuse of auxiliary tasks may significantly contribute to negative transfer.

Combining Tables \ref{tab:structural_metrics} and \ref{algos_ability}, we find that, on the one hand, for different task sets, the same algorithm has different optimal density preferences within the range of optional densities. In other words, the frequency of knowledge transfer is inextricably linked to the characteristics of the task set itself and cannot be directly determined by a certain density to calculate the optimal knowledge transfer parameters. A good EMaTO algorithm should have the ability to adaptively adjust the frequency of knowledge transfer occurrences. On the other hand, even for the same problem, heterogeneous transfer mechanisms require adaptation to different frequencies of transfer behavior because the effectiveness of transfer mechanisms is influenced by specific situations.

\section{Conclusion}
This paper investigates knowledge transfer in many-task evolutionary optimization algorithms through the lens of complex network analysis. Our findings validate the use of directed networks to effectively model task interrelations in these algorithms. We outlined methods to extract knowledge transfer networks from existing algorithms and analyzed them using key network metrics. Our analysis of these networks, particularly from algorithms that performed well on test sets, revealed three salient features: optimal network density, a heterogeneous degree distribution, and pronounced community structures. Notably, knowledge transfer across different communities can result in negative transfer. For future research, integrating core complex network techniques like community detection may enhance the adaptability of knowledge transfer mechanisms in these algorithms.

\bibliographystyle{IEEEtran}
\bibliography{main}
\end{document}